\documentclass[runningheads]{llncs}
\usepackage{graphicx}
\usepackage{amsmath}
\usepackage{amssymb}

\begin{document}

\title{A PCA based Keypoint Tracking Approach to Automated Facial Expressions Encoding}

\author{Shivansh Chandra Tripathi \and
Rahul Garg}

\authorrunning{S. Tripathi and R. Garg}

\institute{Indian Institute of Technology Delhi, New Delhi, India \\
\email{\{shivansh, rahulgarg\}@cse.iitd.ac.in}}

\maketitle              
\begin{abstract}
 The Facial Action Coding System (FACS) for studying facial expressions is manual and requires significant effort and expertise. This paper explores the use of automated techniques to generate Action Units (AUs) for studying facial expressions. We propose an unsupervised approach based on Principal Component Analysis (PCA) and facial keypoint tracking to generate data-driven AUs called PCA AUs using the publicly available DISFA dataset. The PCA AUs comply with the direction of facial muscle movements and are capable of explaining over 92.83 percent of the variance in other public test datasets (BP4D-Spontaneous and CK+), indicating their capability to generalize facial expressions. The PCA AUs are also comparable to a keypoint-based equivalence of FACS AUs in terms of variance explained on the test datasets. In conclusion, our research demonstrates the potential of automated techniques to be an alternative to manual FACS labeling which could lead to efficient real-time analysis of facial expressions in psychology and related fields. To promote further research, we have made code repository publicly available.\footnote{The code can be found here: \url{https://github.com/Shivansh-ct/PCA-AUs}}

\end{abstract}

\section{Introduction} Facial expressions are important for non-verbal communication in humans. They are one of the reliable indicators of vital emotions such as happiness, fear, disgust, anger, and surprise. Owing to their importance in emotion analysis, facial expressions have a wide range of applications in marketing, healthcare, cinematics, security, etc., and are extensively investigated in affective computing~\cite{zhi2020comprehensive}. The most comprehensive and descriptive tool to manually study facial expressions is the Facial Action Coding System (FACS)~\cite{ekman2002facial}.
Facial expressions arise due to the anatomical pull of the facial muscles in different directions. FACS groups the most granular movement on the face produced by a facial muscle or a few of their combinations as Action Units (AUs). Any facial expression can be encoded in terms of AUs, establishing the universality of FACS. Experts trained to annotate facial expressions with AUs are called FACS coders.
\par However, FACS coding is extremely time-intensive~\cite{bartlett1999measuring} limiting its applications in real-time systems such as online marketing and healthcare. A FACS coder requires approximately 100 hours of training before certifying to code AUs~\cite{harrigan2008new}. Also, different coders can interpret facial expressions slightly differently which can lead to inconsistency in the data. Therefore, to eliminate manual dependency various automated systems for AU and emotion analysis are developed using machine learning~\cite{lekshmi2008analysis,de2011facial,mahoor2011facial,mistry2013literature,meher2014face,zhao2016joint,patil2016performance,shao2018deep,zhi2020comprehensive}. Most of these techniques are supervised in nature~\cite{mahoor2011facial,zhao2016joint,shao2018deep,zhi2020comprehensive} and rely heavily on large annotated datasets of AUs which again imposes a dependency on manual FACS coding. Unsupervised techniques~\cite{lekshmi2008analysis,de2011facial,mistry2013literature,meher2014face,patil2016performance,zhi2020comprehensive,sariyanidi2014automatic} have been utilized in facial expression research and can eliminate the need for manual labeling. However, the prime focus of these unsupervised techniques was feature generation to improve AU and emotion recognition rather than the development of data-driven AUs that comply with facial muscle movements. Also, the ability of the generated features to code multiple datasets to establish their universality like FACS AUs stands unexamined.
\par We propose an unsupervised learning method based on Principal Component Analysis (PCA) for facial expression coding. Results show that the PCA components termed PCA AUs can explain large variances in different datasets. The PCA AUs in which the facial keypoint movements are in accordance with the direction of facial muscle movement are defined as interpretable. We show that 50 percent of the PCA AUs are interpretable. These AUs also explain variances in the test datasets comparable to a keypoint-based equivalence of FACS AUs.
\par The next section describes the preprocessing of the facial keypoints and building the PCA model. Section 3 discusses the experimentation and results, and section 4 concludes with the limitations and future scope of our work.

\section{Facial Keypoint-Based Automated Coding System}
In order to track facial expressions in a video sequence, certain keypoints in various points of the face (such as eyes, eyebrows, nose, lips and jawline) are labelled and tracked. The labelling of these keypoints can be done manually or using automated algorithms~\cite{dong2020supervision}. For our analysis, we use the groundtruth facial keypoints provided in the following three publicly available datasets- DISFA~\cite{mavadati2013disfa}(66 keypoints), BP4D-Spontaneous~\cite{zhang2014bp4d}(49 keypoints) and CK+~\cite{lucey2010extended}(68 keypoints). To maintain consistent representation, all the datasets are vectorized to 68 keypoint template (Fig.~\ref{datasetskp}\footnote{Modification of the keypoint image from https://github.com/Fang-Haoshu/Halpe-FullBody/blob/master/docs/face.jpg}), filling the keypoint indices with zero values that are absent in the corresponding data. We preprocess these keypoints to eliminate geometric variabilities across multiple subjects such as head movement, the difference in face sizes, or relative position of face parts. These keypoints are finally converted into features that are representative of facial expressions. Further, we present our PCA model and prepare a keypoint-based equivalence of the FACS AUs. We first discuss the geometric corrections in detail:

\subsection{Geometric Corrections}There are three steps in eliminating the geometric variabilities, frontalization, affine registration and similarity registration as described below:
\par To ensure the presence of pure facial muscle movements, we remove any head movement in our analysis and bring the keypoint to a front-facing image using the algorithm mentioned in Vonikakis et al.~\cite{9190989}.
\par
\begin{figure}[t!]
    \centering
\includegraphics[scale=0.10]{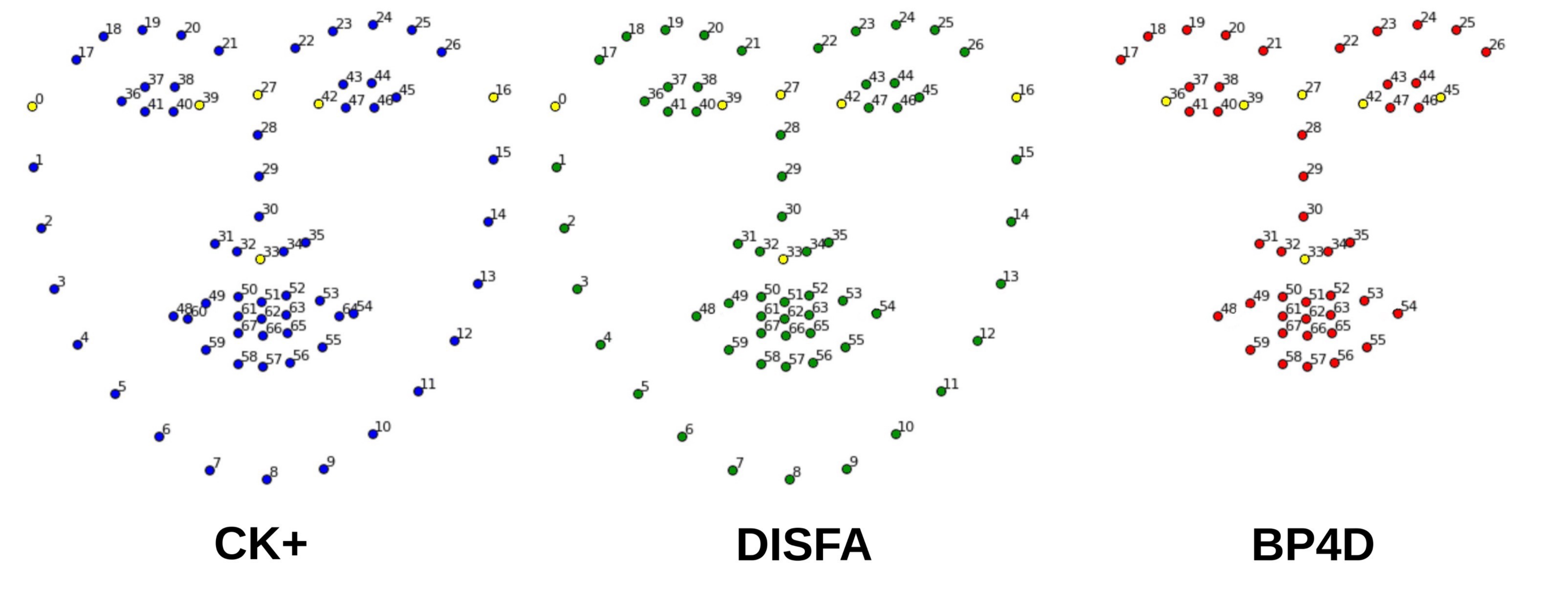}
    \caption{Keypoint locations for the 68-keypoint template in the three datasets. Yellow keypoints indicate the standard affine co-ordinates}
    \label{datasetskp}
\end{figure}

 Next, an affine registration~\cite{hartley2003multiple} registers the keypoints to a standard space to remove face size and position-related variabilities. A set of fixated six facial keypoints are chosen for estimating the geometric parameters in a face image and the remaining keypoints in that image are registered using those parameters. For DISFA and CK+, the six keypoints are: 0,16,39,42,27,33 (Fig.~\ref{datasetskp}). BP4D-Spontaneous has the jawline keypoints (0,16) missing, therefore we use the following keypoints: 39,42,36,45,27,33 (Fig.~\ref{datasetskp}). 

\par Lastly, we eliminate variabilities such as the distance between the eye corners, the length of the nose and eyebrows, and so on using similarity registeration~\cite{hartley2003multiple}. The following fixated keypoints are used for registration: (42,45) keypoints for the left eyebrow and left eye, (36,39) for the right eyebrow and right eye, (27) for the nose, and (0,16) for the jawline. For example, we estimate similarity parameters using left eye corners (42,45), and the remaining left eyebrow and left eye keypoints are registered using those parameters. Lips have no fixated points as all of them move significantly, e.g., in a smile, therefore, we do not perform any similarity registration for the lips.

\subsection{Feature Generation} A feature is defined as a vector representing changes in facial keypoint positions from a neutral face. To generate features, we subtract the neutral frame keypoint values in the x and y directions with the corresponding values in all other frames of the same subject. The data from all subjects are compactly represented in a matrix $X\in\mathbb{R}^{136\times m}$ where $m$ represents the total number of frames in the corresponding dataset.

\subsection{Principal Component Analysis (PCA)} In facial expression research, PCA is widely used for feature extraction and analyzing different facial representations~\cite{lekshmi2008analysis,meher2014face,patil2016performance}. PCA is a classical example of a dimensionality reduction problem and aims to find a lower dimensional representation of data while retaining maximum information. Mathematically, given, $X\in\mathbb{R}^{p\times m}$ where each sample $x_{i}$ is represented in a $p$-dimensional space, PCA finds a set of $k$ basis vectors $u_{i}\in\mathbb{R}^{p}(i=1,2,...,k)$ which can represent the samples in a much lower dimensional space such that $k<<p$. In matrix notation, the PCA decomposition is written as $X\approx UDV$ where $U\in \mathbb{R}^{p\times k}$, $D\in \mathbb{R}^{k\times k}$ and $V\in \mathbb{R}^{k\times m}$ and, $D$ is a diagonal matrix, $U$ and $V$ are orthonormal matrices. $D$ can be absorbed in matrix $V$ as $DV\rightarrow V$ and the decomposition can simply be stated as $X\approx UV$. Any $x_{i}$ is thus a linear combination of the $k$ $u_{i}$'s as $x_{i}\approx\sum_{j=1}^{k}u_{j}v_{ij}$, linear weights being given by matrix $V$. The columns of $U$ are called principal components and $V$ is called the weight matrix. The $k$ basis vectors represent facial keypoint movements from a neutral face. Since only $k$ basis vectors can represent the entire dataset $X$, they are similar to FACS AUs that can encode any facial expression excluding head movements using only 26 AUs. We call the basis vectors in the matrix $U$ as PCA AUs.\\\\
\textbf{Computing weight matrix for a test data: }This step shows how a test data $Y\in \mathbb{R}^{p\times n}$ is represented in terms of the PCA AUs such that $Y\approx \hat{Y}=UV{'}$ or equivalently $\hat{y}_{i}=\sum_{j=1}^{k}u_{j}v_{ij}^{'}$. Since, $U$ is an orthonormal matrix, $V^{'}$ can be computed by projecting $Y$ onto the columns of matrix $U$ as $U^{T}Y$. So, $\hat{Y}=UU^{T}Y$.\\\\
\textbf{Handling keypoint disparity between the datasets: }Since, all the datasets are vectorized to 68 keypoints, any PCA component will be a 136-dimensional vector. We define the keypoints that are originally present in any dataset as non-redundant and redundant otherwise. Therefore, in any PCA component, the number of non-redundant keypoints will be 66,49,68 for the three training datasets- DISFA, BP4D-Spontaneous and CK+ respectively. A PCA model trained on a given dataset can only reconstruct those keypoints in test data that are non-redundant in its training dataset. The following problems can arise due to the disparity in the number of non-redundant keypoints:
\begin{enumerate}
    \item Train contains less non-redundant keypoints than test - In this case, some of the non-redundant keypoints in the test data may not be reconstructed. For, e.g., a PCA trained on BP4D-Spontaneous can only reconstruct 49 keypoints (see BP4D-Spontaneous in Fig.~\ref{datasetskp}) on a DISFA test data. The jawline keypoints in DISFA would not be reconstructed.
    \item Train contains more non-redundant keypoints than test - In this case, some of the redundant keypoints in test may get reconstruced if they are non-redundant in train. For. e.g., a PCA trained on CK+ can reconstruct all 68 keypoints, but if the test data is BP4D-Spontaneous in this case, then the jawline points are also reconstructed in test which are redundant.
\end{enumerate}

\par The above cases can add errors in the performance evaluation. To avoid this, only the non-redundant keypoints that are common between train and test data are kept in the PCA components $U$ derived from train and test data $Y$ when computing weight matrix for $\hat{Y}$. Also, only these keypoints are kept in test data $Y$ and $\hat{Y}$ when we compute the evaluation metrics for performance.

\subsection{Feature Generation for FACS AUs} To compare the PCA AUs with FACS AUs, we convert the FACS AUs to equivalent facial keypoint movements. First, we select a single subject and extract its image for the APEX frame of each AU and AU combination that excludes head movement from the images used in FACS manual. A neutral image of the same subject is also selected. We generate 68 facial keypoints on each of these images using a keypoint tracking algorithm~\cite{dong2020supervision}. Next, these keypoints undergo the same preprocessing steps as mentioned in sections 2.1 and 2.2, to give a 136 dimensional feature vector representing the keypoint motion from neutral to APEX frame. We compile the preprocessed data for 26 pure AUs in a matrix $U_{AU}\in\mathbb{R}^{136\times 26}$. Data for pure AUs along with their various combinations (total 113 in numbers) given in FACS manual are compiled in a matrix $U_{AUC}\in\mathbb{R}^{136\times 113}$ and are called comb AUs.\\

\textbf{Computing weight matrix for PCA vs. FACS comparison: }
This step shows how a test data $Y\in \mathbb{R}^{p\times n}$ is represented in terms of the PCA AUs and FACS AUs
such that $Y\approx UV^{'}$, $Y\approx U_{AU}V^{'}$ or $Y\approx U_{AUC}V^{'}$. PCA gives an ordered list of components in decreasing order of importance. However, there is no such ordering for FACS AUs. For a fair comparison between PCA AUs and FACS AUs, we use lasso for feature selection as well as regression for computing the weight matrix for these AUs.
We employ the lasso modification to LARS, the LARS-EN algorithm~\cite{zou2005regularization} for this purpose and solve the lasso regression objective with a least angle regression to learn $V^{'}$.
The lasso regression objective is given by: $\min \|y_{i}-Uv^{'}_{i}\|^{2}_{2}+\alpha\|v^{'}_{i}\|_{1}$ where $y_{i}$ and $v_{i}$ are columns of Y,V, $\|\|_{2}$ is L2-norm and $\|\|_{1}$ is L1-norm. To limit the number of components used by $y_{i}$, the number of non-zero weights in $v^{'}_{i}$ can be constrained using early stop on the number of LARS-EN algorithmic steps~\cite{zou2005regularization}.

\par pure AUs and comb AUs have 68 non-redundant keypoints. The non-redundant keypoints for PCA AUs will depend on its training data. The keypoint disparity between any of the AU type (PCA AUs, pure AUs or comb AUs) and the test data while computing the weight matrix and the evaluation metrics is handled in the same manner as done in section 2.3.

\section{Experiments}We first outline the datasets and the evaluation metrics used. We then investigate the PCA AUs generated by these datasets for their coding power and interpretability. We finally compare the PCA AUs with the keypoint-based equivalence of the FACS AUs.

    \subsection{Datasets} We used the datasets CK+~\cite{lucey2010extended}, DISFA~\cite{mavadati2013disfa} and BP4D-Spontaneous~\cite{zhang2014bp4d} to evaluate our PCA model. Below we describe the details:\\\\
    \textbf{CK+:} This dataset contains posed facial expressions of 123 subjects and consists of 10,100 video frames. The keypoint movement features extracted from this data on 10,100 frames is called CK+.\\\\
    \textbf{DISFA:} This dataset contains spontaneous facial expressions recording of 27 subjects and consists of a total of nearly 1,14,000 video frames. The keypoint movement features extracted on all these frames is called DISFA\_full. To have a fair comparison with a model trained on CK+, we randomly extract features on same number of frames as CK+, i.e., on 10,100 frames from DISFA\_full to use for training and call it DISFA\_train.\\\\
    \textbf{BP4D-Spontaneous:} This dataset contains spontaneous facial expressions recording of 41 subjects and comprises nearly 1,36,000 video frames. The keypoint movement features extracted on all the frames is called BP4D\_full. Again, for fair comparison, we randomly extract features on 10,100 frames from BP4D\_full to use for training and call it BP4D\_train.

    \subsection{Evaluation Metrics} We used the following three evaluation metrics:
        \begin{enumerate}
            \item \textbf{Train Variance Explained (Train VE) - } Let $X, X\in \mathbb{R}^{p\times m}$ be the training data of a model and $\hat{X}, \hat{X}\in \mathbb{R}^{p\times m}$ is computed using the model parameters such that $\hat{X}=UV$, then Train VE is equal to $100(1-\frac{\| X-\hat{X}\|_{F}^{2}}{\| X\|_{F}^{2}})$.\\
            \item \textbf{Test Variance Explained (Test VE) - } In the case of test data $Y$ and its approximation $\hat{Y}=UV^{'}$, we define Test VE as $100(1-\frac{\| Y-\hat{Y}\|_{F}^{2}}{\| Y\|_{F}^{2}})$ for test time performance.
            \item \textbf{Mean Components (MC) - } MC is the average number of components used by a sample in a dataset. Formally, if $X\in \mathbb{R}^{p\times m}$ is the dataset approximated as $\hat{X}=UV$, then $MC(\hat{X})$ is equal to $\frac{\sum^{i=m}_{i=1}\|v_{i}\|_{0}}{m}$, where $\|v_{i}\|_{0}$ is the L0-norm of $v_{i}$. For a test data $Y\approx \hat{Y}=UV^{'}$, $MC(\hat{Y})$ can be computed in a similar manner.
        \end{enumerate}

\subsection{PCA AUs}

\begin{figure}[t!]
    \centering
\includegraphics[scale=0.14]{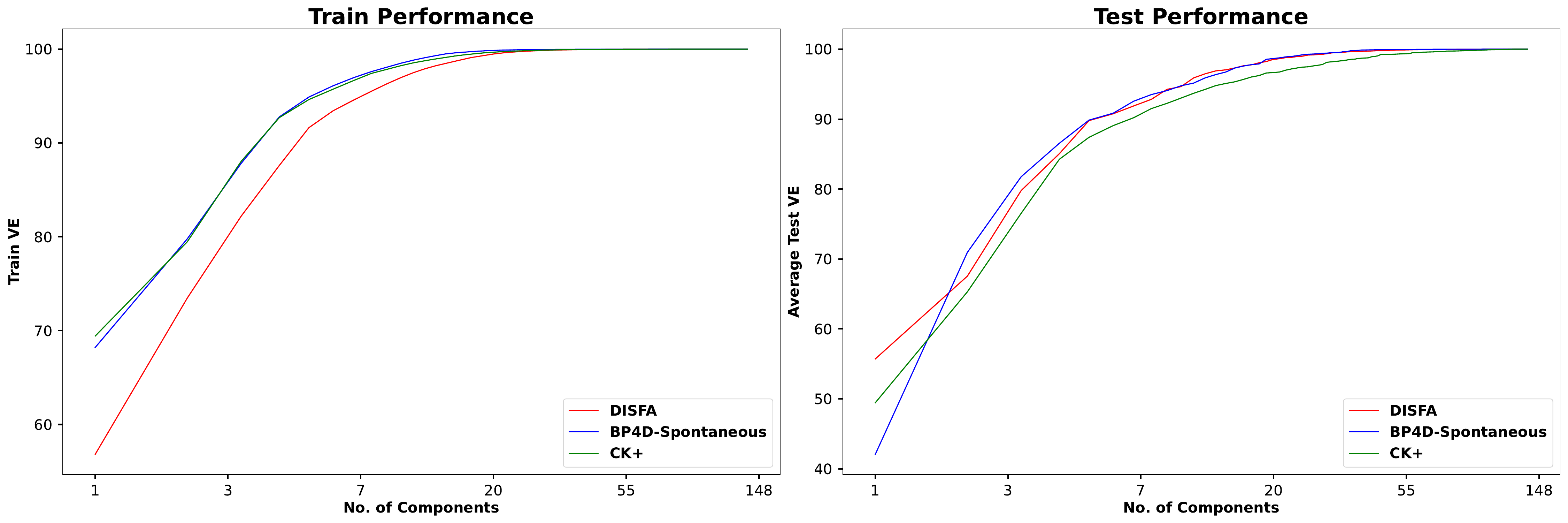}
    \caption{Train VE and the average of the Test VE on remaining two datasets when PCA is trained on either DISFA, CK+ or BP4D-Spontaneous with $k=1$ to $k=136$ (x-axis on log scale)}
    \label{pcaperformance}
\end{figure}

To train a PCA algorithm on a dataset $X$, we tune the hyperparameter $k$ representing the number of components from $k=1$ to $k=136$. For a given train dataset, at each $k$, we compute the Train VE and the average of the Test VE on the remaining two datasets and plot it (Fig.~\ref{pcaperformance}). Average Test VE of DISFA\_train and BP4D\_train are comparable while both of them have consistently better average Test VE than CK+. Table~\ref{pcavar} shows that for a PCA trained with 95 percent Train VE on the three datasets, DISFA\_train outperforms the other two datasets although having comparable performance with BP4D\_train. Finally, although DISFA\_train and BP4D\_train have comparable performance, the PCA trained on DISFA\_train is selected as the final PCA AUs, since it makes a better visualization of facial keypoint movements compared to BP4D\_train which lacks the entire jawline.

\begin{table}[b!]\centering
\caption{Variance Explained (VE) results for PCA}\label{pcavar}
\scriptsize
\begin{tabular}{lrrrrrr}\hline
\textbf{Train Data} & \textbf{Test Data} & \textbf{Train VE} & \textbf{Test VE}\\\hline
DISFA\_train &BP4D\_full &95.51 &90.76 \\
DISFA\_train &CK+ &95.51 &94.90 \\\hline
\multicolumn{2}{c}{\textbf{Average}} &\textbf{95.51} &\textbf{92.83} \\\hline
BP4D\_train &DISFA\_full &96.10 &88.72 \\
BP4D\_train &CK+ &96.10 &92.97 \\\hline
\multicolumn{2}{c}{\textbf{Average}} &\textbf{96.10} &\textbf{90.85} \\\hline
CK+ &DISFA\_full &95.74 &90.06 \\
CK+ &BP4D\_full &95.74 &88.08 \\\hline
\multicolumn{2}{c}{\textbf{Average}} &\textbf{95.74} &\textbf{89.07} \\\hline
\end{tabular}
\end{table}

\setlength{\belowcaptionskip}{-7pt}

\begin{figure}[t!]
    \centering
\includegraphics[scale=0.07]{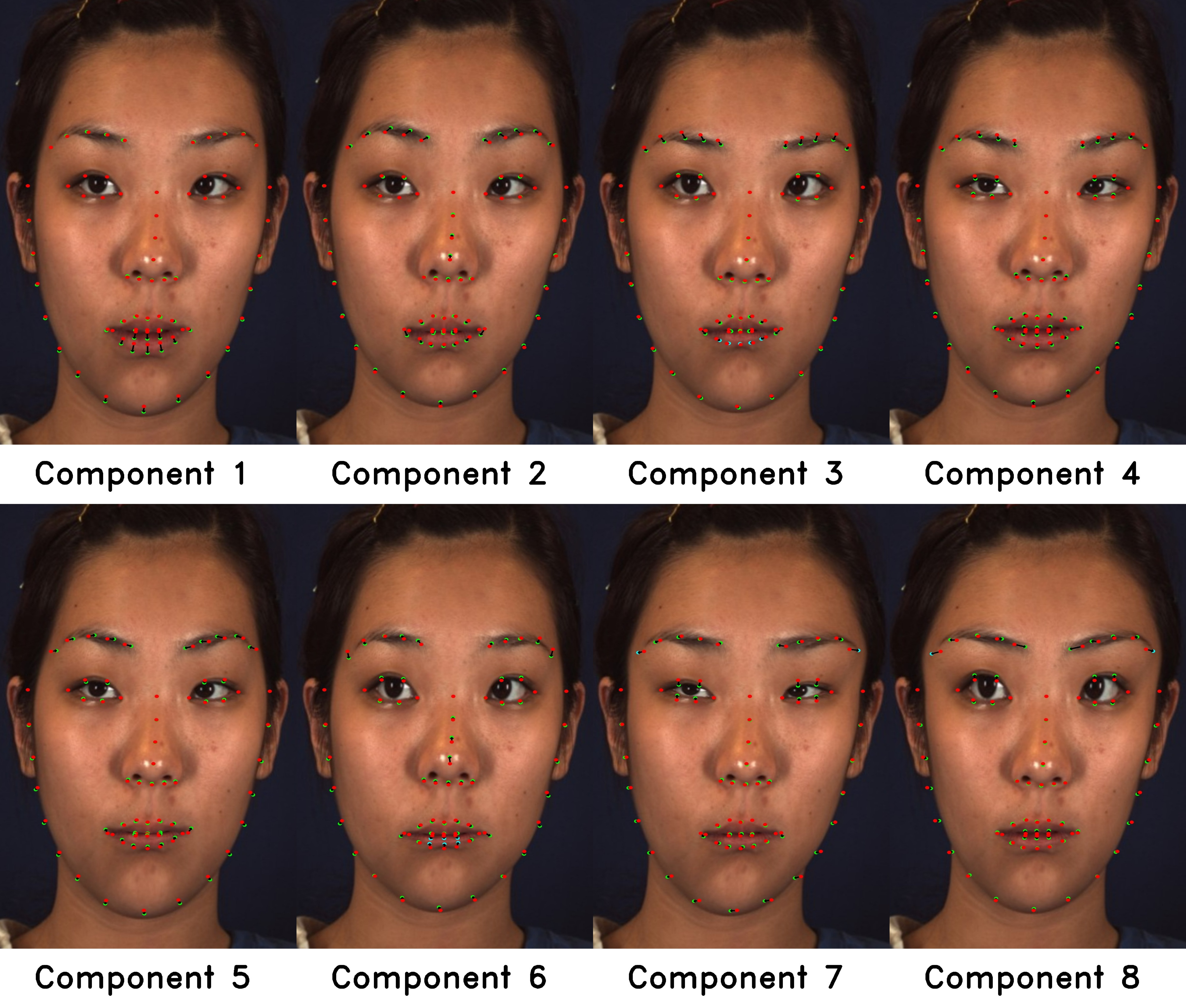}
    \caption{PCA AUs- the red color represents the keypoint position on a neutral face, green on an expression face and blue is a non-interpretable movement. The arrow shows the direction of movement of keypoints. (Face image source: BP4D-Spontaneous~\cite{zhang2014bp4d})}
    \label{pcaau}
\end{figure}

\begin{figure}[b!]
    \centering
\includegraphics[scale=0.08]{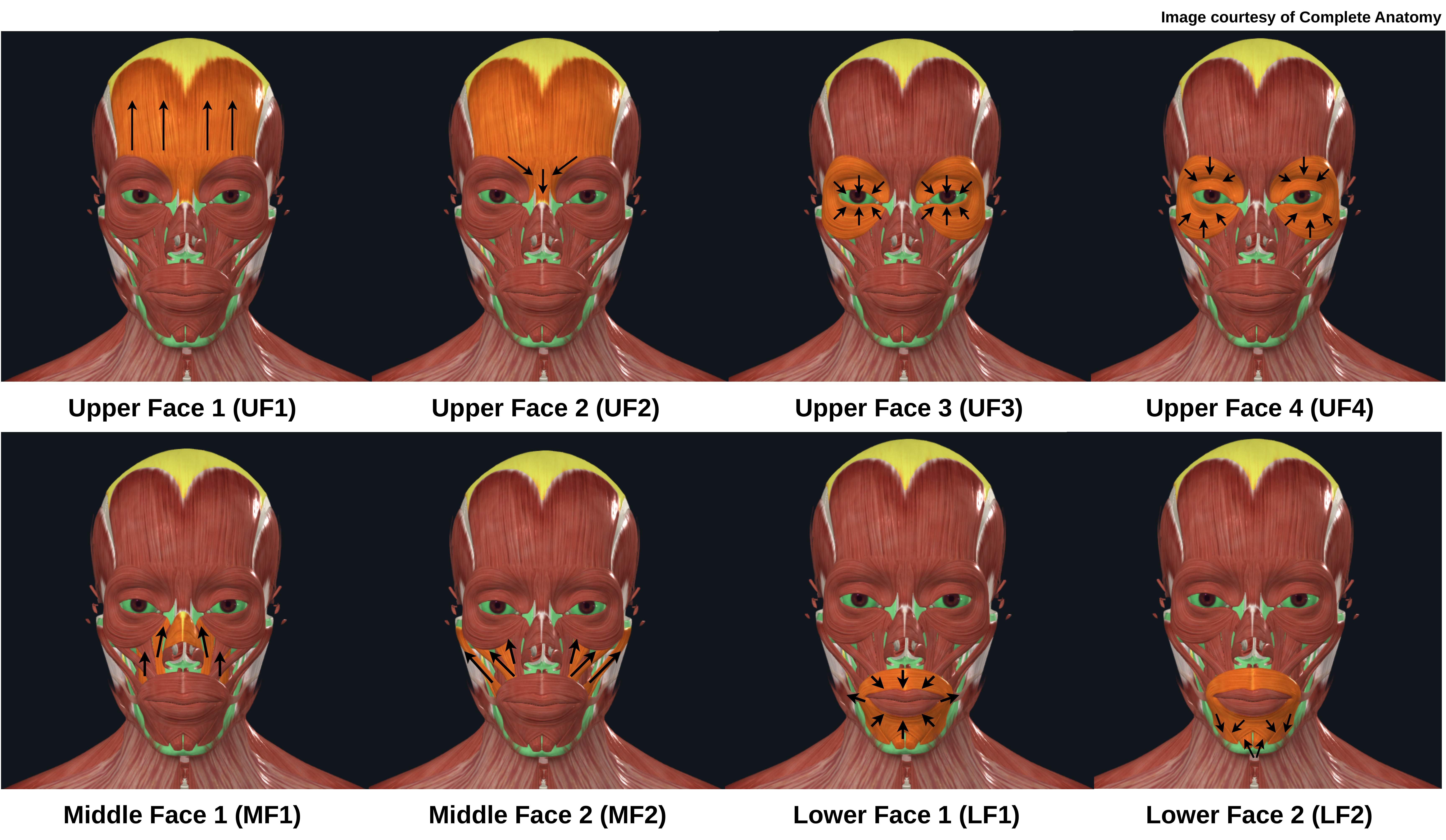}
    \caption{Muscle movements generating AUs that exclude head movements as given by FACS}
    \label{musclefig}
\end{figure}

\par We further investigate whether the keypoints movements in PCA AUs are interpretable by picking up the first eight components that account for 95 percent Train VE. Fig.~\ref{pcaau} shows the visualization of these components in PCA AUs when projected on a neutral face. Fig.~\ref{musclefig} shows major facial muscles responsible for AUs according to FACS that do not include head movements. In PCA component 3, the lower lip move along the line of lower lip towards the centre as shown by blue dots in Fig.~\ref{pcaau} and in PCA component 6, the lower and upper part of lip compress against each other. LF1 and LF2 show that the lower lip is pushed radially towards the centre, or pulled sideways or downwards but not unlike in component 3 and component 6. Also, UF1, UF2, UF3 and UF4 shows that eyebrows are either pulled upward, diagonally towards nose, towards eye socket, but not diagonally down towards the ears as occuring in component 7,8 (Fig.~\ref{pcaau}).

We conclude that components 3,6,7, and 8 contain movements that are non-interpretable. Therefore, among the top eight components of PCA AUs, 50 percent are interpretable and can account for 92.83 percent variance explained in the test datasets on average.

\subsection{Comparison of PCA AUs with FACS AUs}
\begin{figure}[t!]
    \centering
\includegraphics[scale=0.14]{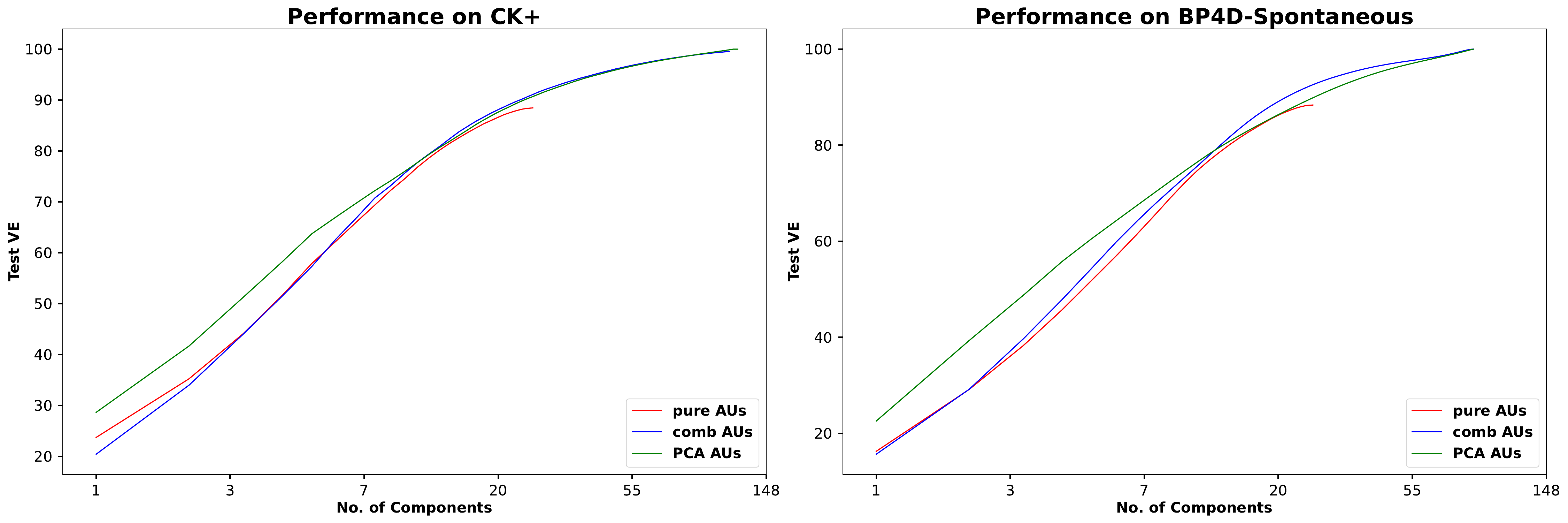}
    \caption{Comparison of PCA AUs, pure AUs and comb AUs (x-axis on log scale)}
    \label{pcavsfacsperformance}
\end{figure}

As mentioned in section 2.4, we use a LARS-EN algorithm to compute the weight matrix for comparing PCA and FACS AUs. On a given test data $Y$ and AU type (PCA AUs, pure AUs or comb AUs), we early stop the LARS-EN algorithm when the average number of components used by $\hat{Y}$ reaches a given value $N_{c}$ such that $MC(\hat{Y})=N_{c}$. We vary $N_{c}$ from 1 to maximum number of AUs present in the given AU type and plot the Test VE for the two datasets- CK+ and BP4D-Spontaneous (Fig.~\ref{pcavsfacsperformance}).
In both CK+ and BP4D-Spontaneous, PCA AUs outperforms FACS AUs when number of components is small. However, asymptotically the PCA AUs and FACS AUs performance becomes comparable. 
\par The keypoint-based FACS AUs are 100 percent interpretable since they are directly extracted from the facial expressions of human images. Overall, we conclude, that the PCA AUs clearly outperforms FACS AUs when $N_{c}<11$ and have a comparable performance for large $N_{c}$. However, they are less interpretable than the keypoint-based FACS AUs.

\section{Conclusion}
We propose an unsupervised technique for facial coding to eliminate the need for manual AU-labelling in datasets which reduces tremendous effort. We find that PCA AUs can code a large number of facial expression samples across datasets and have a reasonable level of interpretability. Our results indicate that the PCA AUs have a comparable ability to code as keypoint-based FACS AUs in terms of variance explained. This approach shows potential for applications in automated emotion analysis and real-time healthcare systems. Future research can focus on increasing the interpretability of the unsupervised generated AUs and validating their performance across a large number of datasets to establish their universality similar to FACS AUs.

\end{document}